\pdfoutput=1
\documentclass[11pt]{article}

\usepackage{emnlp2021}

\usepackage{times}
\usepackage{latexsym}
\usepackage{xcolor}
\usepackage{booktabs}
\usepackage{amsmath}
\usepackage{amssymb}
\usepackage{bbm}
\usepackage{enumitem}
\usepackage{algpseudocode}
\usepackage[noend,ruled,linesnumbered]{algorithm2e}
\usepackage{array,multirow,graphicx}
\usepackage{float}
\usepackage{subfigure}
\usepackage{longtable}
\usepackage{supertabular}
\newcommand{\bs}[1]{\boldsymbol{#1}}
\newcommand{\ie}{\emph{i.e.}}
\newcommand{\eg}{\emph{e.g.}}
\newcommand{\roster}{\textbf{RoSTER}\xspace}

\usepackage[T1]{fontenc}
\usepackage[utf8]{inputenc}

\usepackage{microtype}

\title{Distantly-Supervised Named Entity Recognition with Noise-Robust Learning and Language Model Augmented Self-Training}

\author{Yu Meng, Yunyi Zhang, Jiaxin Huang, Xuan Wang, \\ \textbf{Yu Zhang, Heng Ji, Jiawei Han} \\
  University of Illinois Urbana-Champaign, IL, USA \\
  \texttt{\{yumeng5, yzhan238, jiaxinh3, xwang174,} \\ \texttt{yuz9, hengji, hanj\}@illinois.edu} \\
  }

\begin{document}
\maketitle

% \showthe\textwidth

\begin{abstract}

We study the problem of training named entity recognition (NER) models using only distantly-labeled data, which can be automatically obtained by matching entity mentions in the raw text with entity types in a knowledge base.
The biggest challenge of distantly-supervised NER is that the distant supervision may induce incomplete and noisy labels, rendering the straightforward application of supervised learning ineffective.
In this paper, we propose (1) a noise-robust learning scheme comprised of a new loss function and a noisy label removal step, for training NER models on distantly-labeled data, and (2) a self-training method that uses contextualized augmentations created by pre-trained language models to improve the generalization ability of the NER model.
On three benchmark datasets, our method achieves superior performance, outperforming existing distantly-supervised NER models by significant margins\footnote{Code can be found at \url{https://github.com/yumeng5/RoSTER}.}.

\end{abstract}

\section{Introduction}

Named entity recognition (NER), which aims at identifying real-world entity mentions (\eg, person, location and organization names) from texts, is a fundamental task in natural language processing with a wide range of applications, including question answering~\cite{khalid2008impact}, knowledge base construction~\cite{etzioni2005unsupervised}, text summarization~\cite{aramaki2009text2table} and dialog systems~\cite{bowden2018slugnerds}. 
In recent years, deep neural models~\cite{Devlin2019BERTPO,huang2015bidirectional,lample2016neural,ma2016end} have achieved enormous success for NER, thanks to their strong representation learning power that accurately captures the entity semantics in textual contexts.
However, a common bottleneck of applying deep learning models is the acquisition of abundant high-quality human annotations, and this is especially the case for training NER models, which require every entity mention to be labeled in a sequence.

\begin{figure}[t]
\centering
\includegraphics[width=1.0\columnwidth]{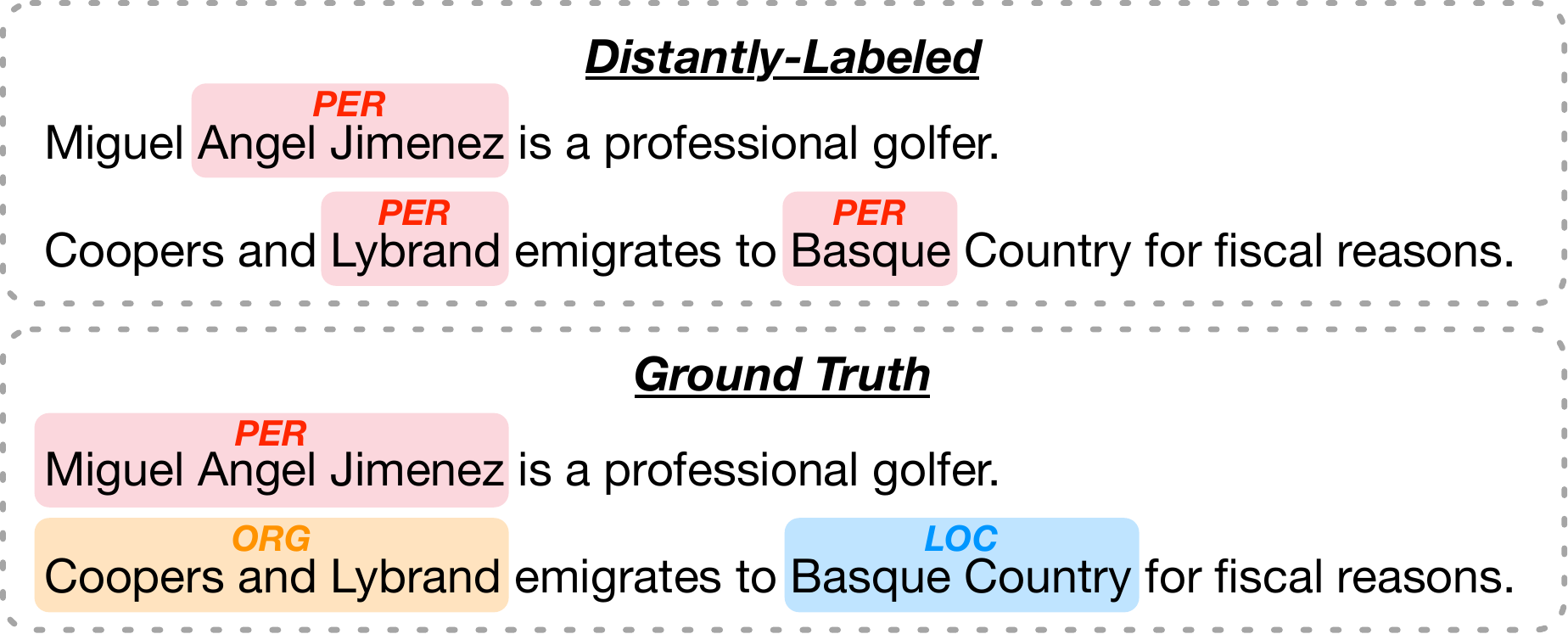}
% \vspace{-1em}
\caption{Distant labels obtained with knowledge bases may be incomplete and noisy, resulting in wrongly-labeled tokens.
% \jh{It seems both are incomplete errors since the 2nd was caused by incompleteness?}
} 
\label{fig:example}
% \vspace{-1.5em}
\end{figure}
To eliminate the need for human annotations, one direction is to use distant supervision for automatic generation of entity labels. 
The common practice is to match entity mentions in the target corpus with typed entities in external gazetteers or knowledge bases.
Unfortunately, such a distant labeling process inevitably introduces incomplete and noisy entity labels, because (1) the distant supervision source has limited coverage of the entity mentions in the target corpus, and (2) some entities can be matched to multiple types in the knowledge bases---such ambiguity cannot be resolved by the context-free matching process.
Figure~\ref{fig:example} shows that some ``\textit{person}'' mentions may be partially labeled (or not labeled at all in other cases), and some entities with multiple possible types may be mislabeled.

Due to the existence of such noise, straightforward application of supervised learning to distantly-labeled data will yield deteriorated performance, because neural models have the strong capacity to fit to the given (noisy) data.
Some previous studies on distantly-supervised NER directly treat distant labels as if ground truth for model training and rely on simple tricks such as applying early stopping~\cite{liang2020bond} and labeling entities with multiple types~\cite{shang2018learning} to handle the noise. Others require an additional manually labeled training set for building a noise classification model~\cite{onoe2019learning}.

In this paper, we study the distantly-supervised NER problem without requiring any human annotations. Our method consists of two steps: (1) noise-robust learning, and (2) language model augmented self-training. 
In the first step, we explicitly address the label noise by using a noise-robust loss function and removing noisy labels.
In the second step, we use the model's high-confidence predictions for self-training to improve generalization, wherein a pre-trained language model is used to not only initialize the NER model, but also generate contextualized augmentations.
Our method is named \roster, for \textbf{Ro}bust learning and \textbf{S}elf-\textbf{T}raining for distantly-supervised \textbf{E}ntity \textbf{R}ecognition.

The contributions of this paper are as follows:
\begin{itemize}[leftmargin=*]
\parskip -0.4ex
\item We propose a noise-robust learning scheme for distantly-supervised NER, comprised of a noise-robust loss function and a noisy label removal step.
\item We propose a new unsupervised contextualized augmentation approach for NER using pre-trained language models. Combined with self-training, the created augmentations improve the model's generalization ability.
\item On three benchmark datasets, \roster outperforms existing distantly-supervised NER approaches by significant margins.
\end{itemize}

% \heng{chagne 'noises' to 'noise' throughout the paper}
% \heng{among these three methods, I found the data augmentation method most exciting, but it's very described in a very abstract way now. Maybe it will be good to be illustrated by some figures and examples. More specifically, I still don't quite get what are the criteria to replace candidates? similarly, in self-training, what are your criteria to choose sentences? only based on confidence values?}

\section{Method}
% In this section, we first briefly describe how to obtain distantly-labeled data using external knowledge bases. 
% Then we introduce three designs to overcome the noise in distant supervision.
% Lastly, we propose to self-train the model using contextualized augmentations produced by pre-trained language models for better generalization.
% At the end of the section, we provide an overall algorithm to summarize our methods.
In this section, we (1) briefly describe how to obtain distantly-labeled data, (2) introduce our noise-robust learning scheme and (3) propose a self-training method with a new contextualized augmentation generation technique.
We assume the pre-trained RoBERTa~\cite{Liu2019RoBERTaAR} model is used as our backbone model, but our proposed methods can be integrated with other architectures (\eg, LSTM-based~\cite{ma2016end}) as well.

\subsection{Distant Label Generation}
\label{sec:dist_label}
Given an unlabeled corpus, the distant labels are usually obtained by matching entities in the corpus with those in the external knowledge bases or gazetteers with typing information.
%JWH  In this work, we do not introduce new distant label generation methods, and follow previous work~\cite{liang2020bond} for this step: 
In this work, instead of introducing new distant label generation methods, we follow the previous work~\cite{liang2020bond} for this step: 
%JWH Firstly, potential entities are determined via POS tagging and hand-crafted rules. Secondly, their types are acquired by querying Wikidata using SPARQL~\cite{vrandevcic2014wikidata}. Lastly, additional gazetteers from multiple online resources are used for matching more entities in the corpus. For more details, please refer to~\cite{liang2020bond}.
(1) potential entities are determined via POS tagging and hand-crafted rules, (2) their types are acquired by querying Wikidata using SPARQL~\cite{vrandevcic2014wikidata}, and (3) additional gazetteers from multiple online resources are used for matching more entities in the corpus.

\subsection{Noise-Robust Learning}
We first overview the common setup for training NER models, and then propose two designs that work jointly for distantly-supervised NER, motivated by the challenges of learning with noisy labels: (1) a new loss function, and (2) noisy label removal. Finally, ensembling multiple models is helpful for stabilizing the model performance.

NER systems are usually trained as a sequence labeling model that classifies every token in a sequence into a set of entity types or non-entity. 
The label space depends on the tagging scheme used (\eg, the BIO format distinguishes begin/inside/outside of named entities). 
Specifically, given a sequence $\bs{x} = [x_1, \dots, x_n]$ of $n$ tokens and their corresponding categorical labels $\bs{y} = [y_1, \dots, y_n]$, an NER model parameterized by $\bs{\theta}$ is trained to minimize some classification loss that encourages the model to correctly predict the given labels.
The cross entropy (CE) loss is most commonly used for such a purpose:
\begin{equation*}
\label{eq:ce_loss}
\mathcal{L}_{\text{CE}} = - \sum_{i=1}^n \log f_{i, y_i}(\bs{x};\bs{\theta}),
\end{equation*}
where $f_{i, j}(\bs{x};\bs{\theta})$ is the model's predicted probability of token $x_i$ belonging to class $j$ (\ie, the softmax layer outputs). 

The gradient of $\mathcal{L}_{\text{CE}}$ is (via the chain rule):
\begin{equation}
\label{eq:ce_grad}
\nabla_{\bs{\theta}} \mathcal{L}_{\text{CE}} = - \sum_{i=1}^n \frac{\nabla_{\bs{\theta}} f_{i, y_i}(\bs{x};\bs{\theta})}{f_{i, y_i}(\bs{x};\bs{\theta})}.
\end{equation}

Due to the $f_{i, y_i}(\bs{x};\bs{\theta})$ term as the denominator, the tokens on which the model's prediction is less congruent with the provided labels (\ie, $f_{i, y_i}(\bs{x};\bs{\theta})$ is smaller) will be implicitly weighed more during the gradient update. Such a mechanism grants better model convergence when trained with clean data (\ie, $\bs{y}$ are ground truth labels), because more emphasis is put on difficult tokens. However, when the labels are noisy, training with the cross entropy loss can cause overfitting to the wrongly-labeled tokens (\eg, the two sentences in Figure~\ref{fig:example}).

Contrary to cross entropy loss which is sensitive to noise, the mean absolute error (MAE) loss, which is commonly used in regression tasks, has been shown inherently noise-tolerant when used for classification~\cite{ghosh2017robust} and is defined as follows (omitting the constant scale factor $2$):
\begin{equation*}
\label{eq:mae_loss}
\mathcal{L}_{\text{MAE}} = \sum_{i=1}^n \left(1 - f_{i, y_i}(\bs{x};\bs{\theta}) \right),
\end{equation*}
and its gradient is given by
\begin{equation}
\label{eq:mae_grad}
\nabla_{\bs{\theta}} \mathcal{L}_{\text{MAE}} = - \sum_{i=1}^n \nabla_{\bs{\theta}} f_{i, y_i}(\bs{x};\bs{\theta}).
\end{equation}

By comparing Eq.~\eqref{eq:mae_grad} with Eq.~\eqref{eq:ce_grad}, we observe that $\mathcal{L}_{\text{MAE}}$ is more noise-robust than $\mathcal{L}_{\text{CE}}$ because Eq.~\eqref{eq:mae_grad} treats every token equally for gradient update, allowing the learning process to be dominated by the correct majority in distant labels. However, using $\mathcal{L}_{\text{MAE}}$ for training deep neural models generally worsens the convergence efficiency and effectiveness due to the inability of adjusting for challenging training samples, and leads to suboptimal model performance compared to using $\mathcal{L}_{\text{CE}}$~\cite{zhang2018generalized}.

\paragraph{Generalized Cross Entropy.}
To balance between model convergence and noise-robustness, we propose to use the generalized cross entropy (GCE) loss~\cite{zhang2018generalized} for training distantly-supervised NER models, inspired by the $q$-order entropy~\cite{ferrari2010maximum}, defined as follows:
\begin{equation}
\label{eq:gce_loss}
\mathcal{L}_{\text{GCE}} = \sum_{i=1}^n \frac{1 - f_{i, y_i}(\bs{x};\bs{\theta})^{q}}{q},
\end{equation}
where $0 < q < 1$ is a hyperparameter: When $q \to 1$, $\mathcal{L}_{\text{GCE}}$ approximates $\mathcal{L}_{\text{MAE}}$; when $q \to 0$, $\mathcal{L}_{\text{GCE}}$ approximates $\mathcal{L}_{\text{CE}}$ (using L’Hôpital’s rule; see Appendix~\ref{sec:gce} for the derivation). The gradient is computed as:
\begin{equation}
\label{eq:gce_grad}
\nabla_{\bs{\theta}} \mathcal{L}_{\text{GCE}} = - \sum_{i=1}^n \frac{\nabla_{\bs{\theta}} f_{i, y_i}(\bs{x};\bs{\theta})}{f_{i, y_i}(\bs{x};\bs{\theta})^{1-q}}.
\end{equation}

Comparing Eq.~\eqref{eq:gce_grad} to Eq.~\eqref{eq:ce_grad}, it can be observed that $\mathcal{L}_{\text{GCE}}$ is more noise-robust than $\mathcal{L}_{\text{CE}}$ because less weights are given to tokens on which the model prediction is less consistent with the given labels (note $f_{i, y_i}(\bs{x};\bs{\theta})^{1-q} > f_{i, y_i}(\bs{x};\bs{\theta})$ for $q > 0$ and $f_{i, y_i}(\bs{x};\bs{\theta}) < 1$). Comparing Eq.~\eqref{eq:gce_grad} to Eq.~\eqref{eq:mae_grad}, it can be seen that $\mathcal{L}_{\text{GCE}}$ facilitates better learning dynamics than $\mathcal{L}_{\text{MAE}}$ because difficult tokens are given more attention to.

\paragraph{Noisy Label Removal.}
Even when a noise-robust loss is used, mislabeled tokens still deteriorate the model performance as long as they are included in training. 
Unfortunately, without any prior knowledge about which tokens are mislabeled, it is challenging to automatically detect them. 
We propose a simple threshold-based strategy to remove noisy labels: At first, all tokens along with their distant labels will be used for model training; later, those tokens on which the model prediction does not strongly agree with its distant label (\ie, $f_{i, y_i}(\bs{x};\bs{\theta}) \le \tau$ where $\tau$ is a threshold value) will be excluded from the training set (\ie, not calculated in the loss). 
The intuition is straightforward: Since our loss function is noise-robust, the learned model will be dominated by the correct majority in the distant labels instead of quickly overfitting to label noise; if the model prediction disagrees with some given labels, they are potentially wrong. 
% \yuz{You may need to explicitly mention that ``excluding'' means skipping them when calculating the loss, not treating them as ``O''.}

Specifically, we extend Eq.~\eqref{eq:gce_loss} to incorporate the aforementioned design, as follows:
\begin{equation}
\label{eq:gce_loss_mod}
\mathcal{L}_{\text{GCE}} = \sum_{i=1}^n w_i \frac{1 - f_{i, y_i}(\bs{x};\bs{\theta})^{q}}{q},
\end{equation}
where $w_i = 1$ at the start of training and is periodically updated once every  several batches as $w_i = \mathbbm{1}\left( f_{i, y_i}(\bs{x};\bs{\theta}) > \tau \right)$, where $\mathbbm{1}(\cdot)$ is the indicator function.

\paragraph{Model Ensemble for Better Stability.}
Due to the stochasticity involved in training neural networks (\eg, dataset shuffling, network random initialization and dropout), models trained with the same algorithm will have different predictions on the same dataset, and this is especially true when inconsistent noisy signals from the distant labels may further disturb model training.
As such, model ensemble is commonly used to suppress the noise and provide better stability, by combining the predictions of multiple models~\cite{laine2016temporal,nguyen2019self}. 
The rationale is that the model’s predictions are likely to be consistent on clean data while inconsistent and oscillating on wrongly-labeled data, and ensembling multiple models enhances consistent predictions and neutralizes inconsistent ones.

We perform model ensemble by simply training $K$ models $\{\bs{\theta}_k\}_{k=1}^K$ via Eq.~\eqref{eq:gce_loss_mod} on the same distantly-labeled corpus with different random seeds controlling the randomness of the training process.
A new model $\bs{\theta}_{\text{ENS}}$ is finally trained to approximate the average prediction of the $K$ models on all tokens by minimizing the Kullback–Leibler (KL) divergence loss:
\begin{equation}
\label{eq:ens_train}
\mathcal{L}_{\text{ENS}} = \sum_{i=1}^n \text{KL} \left( \bar{f}_{i}\left( \bs{x}; \{\bs{\theta}_k\}_{k=1}^K \right) \| f_{i}(\bs{x};\bs{\theta}_{\text{ENS}}) \right),
\end{equation}
where $\bar{f}_{i}\left( \bs{x}; \{\bs{\theta}_k\}_{k=1}^K \right) = \frac{1}{K}\sum_{k=1}^K f_{i}(\bs{x};\bs{\theta}_k)$ is the $K$ models' averaged prediction, and we find that $K = 5$ is sufficient to provide stable ensembled model performance.

% \paragraph{Discussions and Remarks.}
% While our methods introduce two additional hyperparameters $q$ and $\tau$ in Eq.~\eqref{eq:gce_loss_mod} compared to supervised NER model training, we will show in Section~\ref{sec:exp} that the performance is rather insensitive to these hyperparameter values within a reasonable range, and that the same set of hyperparameters can be used for different datasets to avoid parameter tuning. 
% % \yuz{Better say ``within a reasonable range''? If the performance is really insensitive to $q$, why do you propose the new loss?}
% For model ensemble, although multiple models need to be trained, the training can be done in parallel and the number of models need not to be very large, so the training time is not significantly longer. Also, model ensemble will not induce higher inference cost after training, as multiple model predictions are ensembled to train one single model in the end, which is the only model needed for inference.

\paragraph{Remarks.}
While our methods introduce two additional hyperparameters $q$ and $\tau$, their values can be kept same for different datasets to avoid parameter tuning. We will also show in Section~\ref{sec:exp} that the model performance is rather insensitive to these hyperparameter values within a reasonable range.
% The multiple Model ensemble will train one single model in the end, which will not induce higher inference cost after training.

\subsection{Language Model Augmented Self-Training}

\begin{figure*}[t]
\centering
\includegraphics[width=1.0\textwidth]{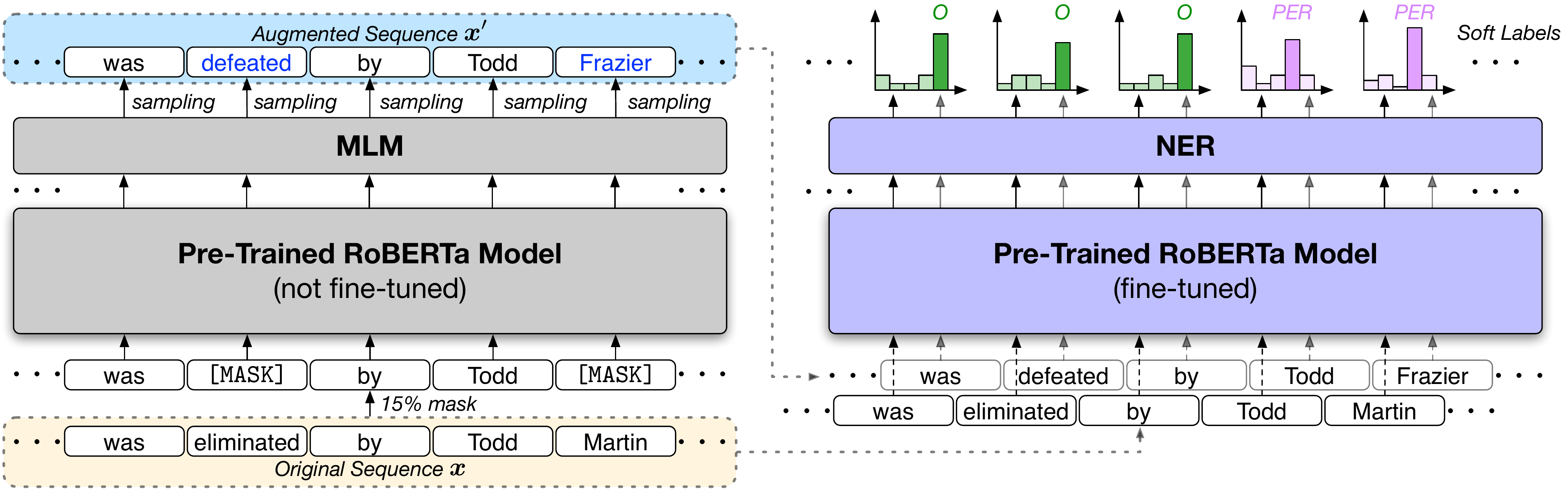}
% \vspace{-1em}
\caption{Overview of language model augmented self-training. Only a part of the sequence is shown; the original sequence is ``Renzo Furlan was eliminated by Todd Martin in the tournament.'' We feed the partially masked original sequence into a pre-trained RoBERTa model and sample from its MLM output probability to obtain an augmented sequence (replaced tokens are marked in blue). Then the NER model is trained with both original and augmented sequences as inputs to approximate the soft labels.} 
\label{fig:aug}
\vspace{-1em}
\end{figure*}

After the noise-robust learning step, we further fine-tune the resulting model (\ie, trained with Eq.~\eqref{eq:ens_train}) via a self-training step on the same corpus, but without the distant labels, for two reasons: (1) The clean signals in the distant supervision have been exploited via noise-robust learning, but some tokens may have not been fully leveraged by the model since they are excluded by the noisy label removal step.
% (especially when a higher threshold value $\tau$ is set for stronger noise-tolerance).
The self-training step aims to bootstrap on all tokens using the model's own predictions to improve its generalization ability.
Similar self-training ideas have been explored in classification tasks~\cite{Meng2018WeaklySupervisedNT,Meng2019WeaklySupervisedHT,meng2020text}.
(2) The pre-trained language model (PLM) has only been used to initialize 
% \yuz{What do you mean by ``initialize''? What more information of PLM is used in self-training?}
the NER model for fine-tuning, while PLMs (without fine-tuning) encode factual and relational knowledge through pre-training~\cite{Petroni2019LanguageMA} that may complement the NER model training. The self-training step thus also brings additional pre-trained knowledge for better model generalization by creating contextualized augmentations using a PLM.
Figure~\ref{fig:aug} shows an overview.

\paragraph{Contextualized Augmentations with PLMs.}
Many PLMs~\cite{Devlin2019BERTPO,Lan2020ALBERTAL,Liu2019RoBERTaAR} are pre-trained with the masked language modeling (MLM) task on large-scale text corpora carrying general knowledge like the Wikipedia.
Previous studies~\cite{jiang2020can,Petroni2019LanguageMA} have shown that entity-related knowledge can be extracted from a PLM (without any fine-tuning) by querying it via cloze templates and gathering the PLM's MLM outputs.

Given that the MLM task shares high similarity with the NER task (\ie, both leverage the contextual information within the sequence for token-level classification) and that the MLM outputs contain general knowledge acquired during pre-training, we propose to use the pre-trained RoBERTa model (without fine-tuning) $\bs{\theta}_{\text{PRE}}$ for creating label-preserving augmentations (\ie, not changing the entity type label or non-entity label) of the original sequences in the corpus in order to complement the NER learning with pre-trained knowledge. Specifically, for each sequence $\bs{x}$ in the corpus, we randomly mask out $15\%$ of its tokens (\ie, replace them with the \texttt{[MASK]} token), and feed the partially masked sequence $\hat{\bs{x}}$ into the pre-trained RoBERTa model. Finally, the augmented sequence $\bs{x}' = [x_1', \dots, x_n']$ is created by sampling from the MLM output probability for each token:
\begin{equation}
\label{eq:aug}
x_i' \sim p_i^{\text{MLM}} \left( \hat{\bs{x}}; \bs{\theta}_{\text{PRE}} \right),
\end{equation}
where $p_i^{\text{MLM}} \left( \hat{\bs{x}}; \bs{\theta}_{\text{PRE}} \right)$ is the MLM probability of the pre-trained RoBERTa model on the $i$th token.

The augmented sequence $\bs{x}'$ will be semantically similar to the original sequence $\bs{x}$, and the replaced tokens in $\bs{x}'$ that are different from those in $\bs{x}$ are likely to be label-preserving
% \footnote{There may be exceptions where the replaced tokens change the entity type of the original tokens, but these cases are uncommon and will not significantly impact training.}
because PLMs are good at predicting missing words in the given context, which are usually interchangeable to the original ones. 
To further enforce the label-preserving constraint of the augmented sequence, we (1) sample $x_i'$ only from the top-$5$ terms given by $p_i^{\text{MLM}} \left( \hat{\bs{x}}; \bs{\theta}_{\text{PRE}} \right)$ to avoid low-quality replacements, and (2) require $x_i'$ to have the same capitalization and tokenization with $x_i$ (\ie, if $x_i$ is capitalized or is a subword, so should $x_i'$).

Using PLMs to perform augmentation for NER has the major benefit of being \emph{unsupervised} and \emph{contextualized}. Without PLMs, one may still perform augmentation by replacing an entity in the sequence with another of the same type in the distant supervision source~\cite{dai2020analysis}. However, such an approach requires prior knowledge about the entity type in the sequence (\ie, it does not work for non-entities or entities not matched with distant labels), and the augmentation is context-free, which may create low-quality and invalid sequences (\eg, it does not fit the context to replace a technology company with a news agency although they both belong to the ``\textit{organization}'' entity type).

\paragraph{Self-Training.}
The goals of self-training (ST) are two-fold: (1) use the model's high-confidence predictions that are likely to be reliable for guiding the model refinement on all tokens, and (2) encourage the model to generate consistent predictions on original sequences and augmented ones, based on the principle that a generalizable model should produce similar predictions for similar inputs. 
To fulfill these goals, we iteratively use the model's current predictions to derive soft labels and gradually update the model so that its predictions on both the original and the augmented sequences approximate the soft labels.

Specifically, at the beginning of self-training, the model $\bs{\theta}^{(0)}$ is initialized to be the model trained with Eq.~\eqref{eq:ens_train}. Then at each iteration $t$, 
% (each iteration consists of $50$ batches)
we derive new soft labels $\bs{y}^{(t+1)}$ that enhance high-confidence predictions while demote low-confidence ones via squaring and normalizing the current predictions on the original sequence $\bs{x}$, following the soft labeling formula by \cite{Xie2016UnsupervisedDE}:
% \vspace{-0.5em}
\begin{align}
\begin{split}
\label{eq:soft_label}
y_{i, j}^{(t+1)} &= \frac{f_{i, j} \left(\bs{x};\bs{\theta}^{(t)} \right)^2 / g_j}{\sum_{j'} \left( f_{i, j} \left(\bs{x};\bs{\theta}^{(t)} \right)^2 / g_{j'} \right)},\\
g_j &= \sum_i f_{i, j}\left(\bs{x};\bs{\theta}^{(t)}\right).
\end{split}
\end{align}
% \vspace{-0.5em}

Then the model $\bs{\theta}^{(t+1)}$ at the next iteration is updated by approximating the soft labels with both the original sequence and the augmented sequence as inputs, via the KL divergence loss:
\begin{align}
\begin{split}
\label{eq:st}
\mathcal{L}_{\text{ST}} &= \sum_{i=1}^{n} \text{KL} \left(y_{i}^{(t+1)} \Big\| f_{i} \left(\bs{x};\bs{\theta}^{(t+1)} \right) \right)\\
&+ \sum_{i=1}^{n} \text{KL} \left(y_{i}^{(t+1)} \Big\| f_{i} \left(\bs{x}';\bs{\theta}^{(t+1)} \right) \right).
\end{split}
\end{align}

Using Eq.~\eqref{eq:st} to train the model not only guides the model learning with its high-confidence predictions, but also encourages consistent predictions between the original and augmented sequences. 
From another perspective, such a training process gradually propagates confident label information from original examples to augmented ones so that the model is trained with more data for better generalization. 

We note that the soft labels in Eq.~\eqref{eq:soft_label} are computed on all entity types, excluding the non-entity class (\ie, the ``O'' class). This is because the ``O'' class usually has many more tokens than any entity type class, while Eq.~\eqref{eq:soft_label} encourages balanced assignments of target soft labels.

\subsection{Overall Algorithm}

We summarize the entire training procedure in Algorithm~\ref{alg:train}. 
Lines $2$-$9$: We train $K$ models with different seeds on the distantly-labeled data using the noise-robust loss. At the end of each iteration, noisy labels are removed based on the model's predictions (Line $9$). 
Line $10$: An ensembled model is trained. 
Line $12$: Contextualized augmentations are created with the pre-trained RoBERTa model. 
Lines $16$-$18$: The soft labels are iteratively computed, and the model is updated to approximate the soft labels on both the original and the augmented sequences.

\begin{algorithm}[ht]
\caption{\roster training.}
\label{alg:train}
\KwIn{
An unlabeled text corpus $\{\bs{x}\}$; external knowledge bases $\Phi$; pre-trained RoBERTa model $\bs{\theta}_{\text{PRE}}$.
}
\KwOut{A trained NER model $\bs{\theta}$.}

$\{\bs{y}\} \gets$ Distant label generation with $\Phi$ \;
// Train $K$ models for ensemble\;
\For{$k \gets 0$ to $K - 1$}
{
    $\bs{\theta}_k \gets \bs{\theta}_{\text{PRE}}$\;
    $\{w_i\}_{i=1}^n \gets 1$\;
    // Train for $M$ iterations\;
    \For{$m \gets 1$ to $M$}
    {
        $\bs{\theta}_k \gets$ Train with Eq.~\eqref{eq:gce_loss_mod}\;
        $\left\{ w_i \gets \mathbbm{1}\left( f_{i, y_i}(\bs{x};\bs{\theta}_k) > \tau \right) \right\}_{i=1}^n$\;
    }
    
}
$\bs{\theta}_{\text{ENS}} \gets$ Train with Eq.~\eqref{eq:ens_train}\;

// Augmentation\;
$\{\bs{x}'\} \gets$ Eq.~\eqref{eq:aug}\;
// Self-training\;
$\bs{\theta}^{(0)} \gets \bs{\theta}_{\text{ENS}}$\;
// Train for $T$ iterations\;
\For{$t \gets 0$ to $T - 1$}
{
    $\bs{y}^{(t+1)} \gets$ Eq.~\eqref{eq:soft_label}\;
    $\bs{\theta}^{(t+1)} \gets$ Train with Eq.~\eqref{eq:st}\;
}
Return $\bs{\theta} = \bs{\theta}^{(T)}$\;
\end{algorithm}
% \vspace{-1em}

\section{Experiments}
\label{sec:exp}

\subsection{Datasets}

\begin{table}[t]
\centering
\begin{tabular}{*{4}{c}}
\toprule
\textbf{Dataset} & \textbf{\# Types} & \textbf{\# Train} & \textbf{\# Test} \\
\midrule
CoNLL03 & 4 & 14,041 & 3,453 \\
OntoNotes5.0 & 18 & 59,924 & 8,262 \\
Wikigold & 4 & 1,142 & 274 \\
\bottomrule
\end{tabular}
% \vspace{-0.5em}
\caption{
Dataset statistics with the number of entity types and the number of training/test sequences. 
% Supervised models are trained on the entire training set. Distantly-supervised models are trained on the distantly-labeled training set. All models are evaluated on the test set.
}
% \vspace{-1em}
\label{tab:dataset}
\end{table}

% \begin{table*}[t]
% \centering
% \begin{tabular}{ll*{10}{c}}
% \toprule
%  & \multirow{2}{*}{\textbf{Methods}} & \multicolumn{3}{c}{CoNLL03} & \multicolumn{3}{c}{OntoNotes5.0} & \multicolumn{3}{c}{Wikigold} \\
% & & \textbf{Pre.} & \textbf{Rec.} & \textbf{F1} & \textbf{Pre.} & \textbf{Rec.} & \textbf{F1} & \textbf{Pre.} & \textbf{Rec.} & \textbf{F1} \\
% \midrule
% \parbox[t]{2mm}{\multirow{5}{*}{\rotatebox[origin=c]{90}{Distant-Sup.}}} 
% % \multirow{5}{*}{\textbf{Weakly-Sup.}}
% & \textbf{Distant Match} & 0.811 & 0.638 & 0.714 & 0.745 & 0.693 & 0.718 & 0.479 & 0.476 & 0.478 \\
% & \textbf{Distant RoBERTa} & 0.837 & 0.633 & 0.721 & 0.760 & 0.715 & 0.737 & 0.603 & 0.532 & 0.565 \\
% & \textbf{AutoNER} & 0.752 & 0.604 & 0.670 & 0.731 & 0.712 & 0.721 & 0.435 & 0.524 & 0.475 \\ 
% & \textbf{BOND} & 0.821 & 0.809 & 0.815 & \textbf{0.774} & 0.701 & 0.736 & 0.534 & 0.686 & 0.600 \\
% & \textbf{\roster (Ours)} & \textbf{0.859} & \textbf{0.849} & \textbf{0.854} & 0.753 & \textbf{0.789} & \textbf{0.771} & \textbf{0.649} & \textbf{0.710} & \textbf{0.678} \\
% \midrule
% \parbox[t]{2mm}{\multirow{2}{*}{\rotatebox[origin=c]{90}{Sup.}}}
% % \multirow{2}{*}{\textbf{Supervised}}
% & \textbf{BiLSTM-CNN-CRF} & 0.914 & 0.911 & 0.912 & 0.888 & 0.887 & 0.887 & 0.554 & 0.543 & 0.549 \\
% & \textbf{RoBERTa} & 0.906 & 0.917 & 0.912 & 0.886 & 0.890 & 0.888 & 0.853 & 0.876 & 0.864 \\
% \bottomrule
% \end{tabular}
% % \vspace{-0.5em}
% \caption{
% Performance of all methods on three datasets measured by precision (Pre.), recall (Rec.) and F1 scores.
% }
% % \vspace{-1em}
% \label{tab:results}
% \end{table*}

\setlength{\tabcolsep}{4.2pt}
\begin{table*}[t]
\centering
\begin{tabular}{ll*{10}{l}}
\toprule
 & \multirow{2}{*}{\textbf{Methods}} & \multicolumn{3}{c}{CoNLL03} & \multicolumn{3}{c}{OntoNotes5.0} & \multicolumn{3}{c}{Wikigold} \\
& & \textbf{Pre.} & \textbf{Rec.} & \textbf{F1} & \textbf{Pre.} & \textbf{Rec.} & \textbf{F1} & \textbf{Pre.} & \textbf{Rec.} & \textbf{F1} \\
\midrule
\parbox[t]{2mm}{\multirow{5}{*}{\rotatebox[origin=c]{90}{Distant-Sup.}}} 
% \multirow{5}{*}{\textbf{Weakly-Sup.}}
& \textbf{Distant Match} & 0.811 & 0.638 & 0.714 & 0.745$^*$ & 0.693$^*$ & 0.718$^*$ & 0.479 & 0.476 & 0.478 \\
& \textbf{Distant RoBERTa} & 0.837$^*$ & 0.633$^*$ & 0.721$^*$ & 0.760$^*$ & 0.715$^*$ & 0.737$^*$ & 0.603$^*$ & 0.532$^*$ & 0.565$^*$ \\
& \textbf{AutoNER} & 0.752 & 0.604 & 0.670 & 0.731$^*$ & 0.712$^*$ & 0.721$^*$ & 0.435 & 0.524 & 0.475 \\ 
& \textbf{BOND} & 0.821 & 0.809 & 0.815 & \textbf{0.774}$^*$ & 0.701$^*$ & 0.736$^*$ & 0.534 & 0.686 & 0.600 \\
& \textbf{\roster (Ours)} & \textbf{0.859} & \textbf{0.849} & \textbf{0.854} & 0.753 & \textbf{0.789} & \textbf{0.771} & \textbf{0.649} & \textbf{0.710} & \textbf{0.678} \\
\midrule
\parbox[t]{2mm}{\multirow{2}{*}{\rotatebox[origin=c]{90}{Sup.}}}
% \multirow{2}{*}{\textbf{Supervised}}
& \textbf{BiLSTM-CNN-CRF} & 0.914 & 0.911 & 0.912 & 0.888$^*$ & 0.887$^*$ & 0.887$^*$ & 0.554 & 0.543 & 0.549 \\
& \textbf{RoBERTa} & 0.906$^*$ & 0.917$^*$ & 0.912$^*$ & 0.886$^*$ & 0.890$^*$ & 0.888$^*$ & 0.853 & 0.876 & 0.864 \\
\bottomrule
\end{tabular}
\vspace{-0.5em}
\caption{
Performance of all methods on three datasets measured by precision (Pre.), recall (Rec.) and F1 scores. Baseline results marked with $^*$ are our own runs; others are reported by \cite{liang2020bond}.
}
% \vspace{-1em}
\label{tab:results}
\end{table*}
\setlength{\tabcolsep}{6pt}
% \begin{table}[t]
% \centering
% \begin{tabular}{l*{3}{c}}
% \toprule
% \textbf{Entity Type} & \textbf{Pre.} & \textbf{Rec.} & \textbf{F1} \\
% \midrule
% % \multirow{5}{*}{\textbf{Weakly-Sup.}}
% Person & 0.919 & 0.950 & 0.934 \\
% Location & 0.861 & 0.870 & 0.865 \\
% Organization & 0.803 & 0.879 & 0.839 \\
% Miscellaneous & 0.860 & 0.497 & 0.630 \\
% \bottomrule
% \end{tabular}
% \vspace{-0.5em}
% \caption{
% Performance of \roster by entity type on CoNLL03.
% }
% \vspace{-1em}
% \label{tab:conll_type}
% \end{table}

We use three benchmark datasets for NER: CoNLL03~\cite{sang2003introduction}, OntoNotes5.0~\cite{weischedel2013ontonotes} which we follow the pre-processing of \cite{chiu2016named}, and Wikigold~\cite{balasuriya2009named}. The dataset statistics are shown in Table~\ref{tab:dataset}. All datasets are in English language.

\subsection{Compared Methods}

We compare with a wide range of state-of-the-art distantly-supervised methods and supervised methods.  Fully supervised methods use the ground truth training set for model training. 
Distantly-supervised methods use the distantly-labeled training set obtained as in \cite{liang2020bond}. 
All methods are evaluated on the test set.

\paragraph{Distantly-supervised methods:}
\begin{itemize}[leftmargin=*]
\item \textbf{Distant Match}:
This is the baseline that reports the distant supervision quality (\ie, compares distantly-labeled results with the ground truth).
\item \textbf{Distant RoBERTa}: 
We fine-tune a pre-trained RoBERTa model on distantly-labeled data as if they are ground truth with the standard supervised learning.
\item \textbf{AutoNER}~\cite{shang2018learning}: 
It trains the neural model with a ``Tie or Break'' tagging scheme. Ambiguous tokens are assigned with all possible labels.
\item \textbf{BOND}~\cite{liang2020bond}: 
It first trains a RoBERTa model on distantly-labeled data with early stopping, and then uses a teacher-student framework to iteratively self-train the model.
\end{itemize}

\paragraph{Supervised methods:}
\begin{itemize}[leftmargin=*]
\item \textbf{BiLSTM-CNN-CRF}~\cite{ma2016end}: 
It was one of the state-of-the-art NER models before the appearance of pre-trained language models, using bidirectional LSTM, CNN and CRF. It is trained from scratch on the training data without any pre-trained knowledge.
\item \textbf{RoBERTa}: 
We fine-tune a pre-trained RoBERTa model on the ground truth training data.
\end{itemize}

\subsection{Experiment Settings}

We use the pre-trained RoBERTa-base model as the backbone model (for our method and baselines). 
For the three datasets CoNLL03, OntoNotes5.0, and Wikigold, the maximum sequence lengths are set to be $150$, $180$, and $120$ tokens. 
For all three datasets: The training batch size is $32$;
the hyperparameters $\tau$ and $q$ used by Eq.~\eqref{eq:gce_loss_mod} are both set as $0.7$;
the number of models for ensemble $K = 5$;
% the number of training epochs $M = 3$;
we use Adam~\cite{Kingma2015AdamAM} as the optimizer. The peak learning rate is $3e-5$, $1e-5$ and $5e-7$ for noise-robust training, ensemble model training and self-training, respectively, with linear decay. 
The model is run on $2$ NVIDIA GeForce GTX 1080 Ti GPUs.
More implementation details can be found at Appendix~\ref{sec:impl}.

\subsection{Main Results}

Table~\ref{tab:results} presents the performance of all methods measured by precision, recall and F1 scores. 
% Our model's performance by entity type on CoNLL03 dataset can be found in Table~\ref{tab:conll_type}.
On all three datasets, \roster achieves the best performance among distantly-supervised methods. 
Specifically, (1) the Distant RoBERTa baseline only slightly improves the distant labeling results, showing that directly applying supervised learning to distantly-labeled data will lead to overfitting to label noise and poor model generalization; (2) \roster consistently outperforms AutoNER and BOND, demonstrating the superiority of our proposed noise-robust learning and self-training approach when trained on distantly-labeled data.

For further comparison with supervised methods, we vary the number of ground truth training sequences used for supervised RoBERTa, and show its performance in Figure~\ref{fig:sup}. The performance of \roster is equivalent to using $1,000$ cleanly annotated sequences for supervised RoBERTa.

\subsection{Ablation Study}

\begin{table}[t]
\centering
\begin{tabular}{l*{3}{c}}
\toprule
\textbf{Ablations} & \textbf{Pre.} & \textbf{Rec.} & \textbf{F1} \\
\midrule
% \multirow{5}{*}{\textbf{Weakly-Sup.}}
\textbf{\roster} & 0.859 & 0.849 & 0.854 \\
\qquad \qquad\textbf{w/o GCE} & 0.817 & 0.843 & 0.830 \\
\qquad \qquad\textbf{w/o NR} & 0.830 & 0.836 & 0.833 \\
\qquad \qquad\textbf{w/o ST} & 0.844 & 0.812 & 0.828 \\
\bottomrule
\end{tabular}
% \vspace{-0.5em}
\caption{
Ablation study on CoNLL03 dataset. We compare our full method with ablations (see texts for the abbreviation meanings).
}
% \vspace{-0.5em}
\label{tab:ablation}
\end{table}

\begin{table}[t]
\centering
\begin{tabular}{lc}
\toprule
\textbf{Ablations} & \textbf{Mean (Std.) F1} \\
\midrule
\textbf{w. ensemble} & 0.828 (0.009) \\
\textbf{w/o ensemble} & 0.817 (0.025) \\
\bottomrule
\end{tabular}
% \vspace{-0.5em}
\caption{
Mean and standard deviation (std.) F1 scores of $5$ runs (before self-training) with and without model ensemble on CoNLL03 dataset.
}
% \vspace{-0.5em}
\label{tab:ensemble}
\end{table}

To further validate the effectiveness of each component, we compare \roster with the following ablations by removing one component at a time: (1) replace the GCE loss in Eq.~\eqref{eq:gce_loss_mod} with cross entropy loss (\textbf{w/o GCE}); (2) do not perform noisy label removal (\textbf{w/o NR}); 
% (3) do not perform model ensemble (\textbf{w/o ENS}); 
(3) do not perform self-training (\textbf{w/o ST}).
The results are shown in Table~\ref{tab:ablation}.
It can be seen that \textbf{w/o GCE} and \textbf{w/o NR} both lead to worse performance than the full model, confirming the necessity of jointly using both designs in noise-robust learning; \textbf{w/o ST} also reduces performance, showing that bootstrapping the model with its own high-confidence predictions benefits the model's generalization.

We also study the effect of model ensemble by running noise-robust training (without subsequent self-training) with $5$ different seeds and reporting the mean and standard deviation F1 in Table~\ref{tab:ensemble}. Ensembling multiple models slightly improves the mean result and greatly reduces the variance.

\subsection{Parameter Study}

\begin{figure*}[t]
% \subfigcapmargin=10pt
\centering
\subfigure[]{
	\label{fig:sup}
	\includegraphics[width = 0.32\textwidth]{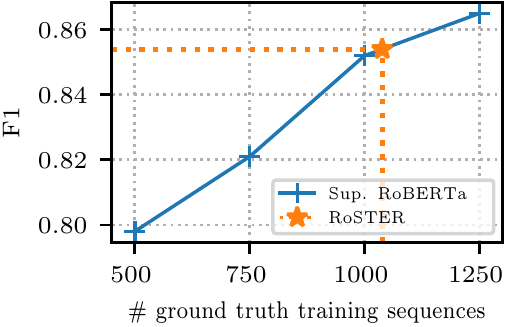}
}
\subfigure[]{
	\label{fig:param}
	\includegraphics[width = 0.315\textwidth]{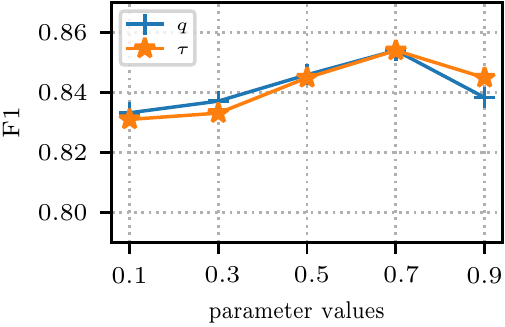}
}
\subfigure[]{
	\label{fig:st}
	\includegraphics[width = 0.315\textwidth]{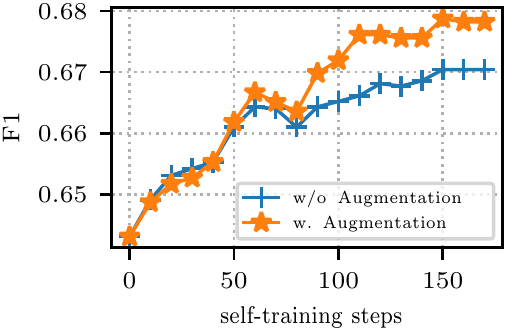}
}
% \vspace{-1em}
\caption{(a) (On CoNLL03) Supervised RoBERTa with different number of ground truth training sequences. \textbf{\roster} trained on distantly-labeled data is equivalent  to supervised RoBERTa using around $1,000$ ground truth sequences. (b) (On CoNLL03) Parameter study. (c) (On Wikigold) Self-training with and without augmentation.
}
% \vspace{-1em}
\label{fig:studies}
\end{figure*}

We study the effect of two important hyperparameters $q$ and $\tau$ used in Eq.~\eqref{eq:gce_loss_mod} on the model performance.
We separately vary the value of $q$ or $\tau$ in range $[0.1, 0.3, 0.5, 0.7, 0.9]$ while keeping the other's value as default (both use $0.7$ as the default value). 
The change in model performance (measured by F1) is shown in Figure~\ref{fig:param}.
Overall, the performance is rather insensitive to the two hyperparameters in the $0.5-0.9$ range. 
When $q \to 1$, $\mathcal{L}_{\text{GCE}}$ approximates $\mathcal{L}_{\text{MAE}}$, having good noise-robustness but poor convergence effectiveness; when $q \to 0$, $\mathcal{L}_{\text{GCE}}$ approximates $\mathcal{L}_{\text{CE}}$, having good convergence but weak noise-robustness. 
Setting $q = 0.7$ achieves a good balance between convergence and noise-robustness.
When $\tau \to 0$, all distant labels will be used for model training, and the model performance will suffer from the noise in them; when $\tau \to 1$, many correct labels will be removed, and there will be insufficient training data. Setting $\tau = 0.7$ allows removing noisy labels while keeping enough reliable training data.

\subsection{Study of Augmentation}

\newcommand{\blue}[1]{\textcolor{blue}{#1}}

\begin{table}[t]
\centering
\begin{tabular}{l}
\toprule
\textbf{Original}: Swiss Bank Corp sets warrants on\\ DTB-Bund-Future. \\
\textbf{Augmentation}: Swiss \blue{Investment} Corp sets\\ warrants \blue{for} \blue{HTB}-Bund-Future. \\
\midrule
\textbf{Original}: Chelsea Clinton was carefully\\ shielded from the exposure of public life. \\
\textbf{Augmentation}: \blue{Hillary} Clinton was \blue{largely}\\ shielded from the \blue{spotlight} of public life. \\
\bottomrule
\end{tabular}
% \vspace{-0.5em}
\caption{
Original sequences and generated augmentations. Replaced words are marked in blue.
}
% \vspace{-1em}
\label{tab:aug}
\end{table}

We study the effectiveness of the generated contextualized augmentations for the self-training step. We run the self-training step with and without using the augmentations (\ie, including or excluding the second term in Eq.~\eqref{eq:st}) on the Wikigold dataset and show the results in Figure~\ref{fig:st}.
Even without augmentations, the self-training improves the model by using high-confidence predictions for self-refinement; with augmentations, the model is trained with more data and eventually generalizes better with higher test set performance. Two concrete augmentation examples are shown in Table~\ref{tab:aug}.

\newcommand{\per}[1]{\textcolor{red}{[#1]$_{\text{PER}}$}}
\newcommand{\org}[1]{\textcolor{orange}{[#1]$_{\text{ORG}}$}}
\newcommand{\loc}[1]{\textcolor{blue}{[#1]$_{\text{LOC}}$}}
\subsection{Case Study}

\begin{table*}[t]
\centering
\scalebox{0.9}{
\begin{tabular}{l}
\toprule
\textbf{Distant Match}: Shanghai-Ek \per{Chor} is jointly owned by the Shanghai Automobile Corporation \\ and \per{Ek Chor} China Motorcycle. \\
% Security forces also killed an unspecified number of members of a rebel gang \\ on Wednesday in the Leveilly suburb of \blue{[Algiers]$_{\text{LOC}}$}, \orange{[Le Matin]$_{\text{PER}}$} newspaper reported. \\
\textbf{Ground Truth}: \org{Shanghai-Ek Chor} is jointly owned by the \org{Shanghai Automobile Corporation}\\ and \org{Ek Chor China Motorcycle}.  \\
\midrule
\textbf{AutoNER}: Shanghai-Ek \per{Chor} is jointly owned by the Shanghai Automobile Corporation \\ and \per{Ek Chor} \loc{China} Motorcycle. \\
\textbf{BOND}: \per{Shanghai-Ek Chor} is jointly owned by the \loc{Shanghai} \org{Automobile Corporation} \\ and \per{Ek Chor} \org{China Motorcycle}. \\
\textbf{\roster}: \org{Shanghai-Ek Chor} is jointly owned by the \org{Shanghai Automobile Corporation}\\ and \org{Ek Chor China Motorcycle}. \\
\bottomrule
\end{tabular}
}
% \vspace{-0.5em}
\caption{
Case study with \roster and baselines. The sentence is from CoNLL03.
}
% \vspace{-1em}
\label{tab:case}
\end{table*}
Finally, we perform case study to understand the advantage of \roster with a concrete example in Table~\ref{tab:case}.
We show the prediction of \textbf{AutoNER}, \textbf{BOND} and \roster on a training sequence with label noise.
\textbf{AutoNER} mainly learns from the given distant labels and slightly generalizes (labels ``China'' separately as a location entity);
\textbf{BOND} is able to generalize better for more complete entity detection because it has a self-training step that bootstraps the model on the training set without completely overfitting to distant labels; however, it is still impacted by label noise.
\roster is able to detect the noisy labels via the noise-robust learning step, and then it further re-estimates the true labels in the self-training step with the help of the reliable signals it learns from the clean data as well as the pre-trained knowledge from PLMs via augmentation, instead of relying purely on distant labels.

\section{Related Work}
% \subsection{Weakly/Distantly-Supervised NER}

% Recent years have witnessed the success of using deep neural models for NER, 
% such as bidirectional LSTM with CRF layers~\cite{ma2016end} and pre-trained language models~\cite{Devlin2019BERTPO,Liu2019RoBERTaAR}, 
% thanks to their strong representation learning power for accurately capturing entity semantics. However, such models' effectiveness comes with the cost of annotating large amounts of training data. 
The effectiveness of deep neural models for NER usually comes with the cost of annotating large amounts of training data. 
To alleviate the human annotation burden when applying deep models, several studies propose to train NER models with weakly/distantly-labeled data. For weakly-supervised NER, previous studies have explored cross lingual knowledge transfer from high resource languages to low resource languages~\cite{feng2018improving,ni2017weakly,xie2018neural}, aggregating multiple weak labeling functions  \cite{lison2020named,safranchik2020weakly} or leveraging sentence-level labels~\cite{kruengkrai2020improving}. Few-shot approaches~\cite{Huang2021FewshotNE} have also been explored to leverage very few labeled data for NER model training.

Our work is more closely related to distantly-supervised NER which uses external gazetteers or knowledge bases to automatically derive entity labels. 
Along this line, different methods have been proposed to leverage the distant supervision, such as propagating reliable type information on graphs~\cite{ren2015clustype}, designing new model components to handle multiple possible labels~\cite{shang2018learning}, employing additional models to classify noisy data~\cite{onoe2019learning}, formulating the task as a positive-unlabeled learning problem~\cite{peng2019distantly}, and adopting early stopping to prevent the model from overfitting to distant labels~\cite{liang2020bond}.
However, previous methods either do not explicitly address the noise in the distantly-labeled data (\ie, treating them as if they are ground truth), or require an additional set of manually-labeled data to train a denoising model.
Our method addresses the label noise with a noise-robust learning scheme and a self-training step for better generalization, without using any ground truth data. 

Our study is also related to data augmentation techniques. In NLP, data augmentation is well developed for text classification, by either creating real text sequences~\cite{xie2020unsupervised} via back translation~\cite{Sennrich2016ImprovingNM} or in the hidden states of the model via perturbations~\cite{Miyato2017AdversarialTM} or interpolations by mixing up labels~\cite{Chen2020MixTextLI}.
However, these techniques cannot be readily used for the NER task.
\cite{dai2020analysis} study a set of simple augmentation methods for the NER task, like synonym replacement, mention replacement or segment shuffling. Nevertheless, these augmentations are context-free which may generate unreasonable sequences or require additional sources like the WordNet. 
% As such, the erroneous semantic/syntactic signals from the augmentations may be harmful for model training. 
Our proposed augmentation method is unsupervised and contextualized, generating high-quality sequences thanks to the pre-trained knowledge of PLMs and reliably improving model generalization.

\section{Conclusion and Future Work}
In this paper, we study the distantly-supervised NER problem without using any human annotations but only distantly-labeled data. For better model training with noisy data, we propose a noise-robust learning scheme, consisting of a new loss function and a noisy label removal step.
To further improve the model generalization, we propose a self-training method that guides model refinement with its own high-confidence predictions and enforces the model to make consistent predictions on original and augmented sequences generated by PLMs.
Our method achieves strong performance on three benchmark datasets, outperforming previous distantly-supervised NER methods.

The techniques proposed in this paper are generalizable for future studies: The noise-robust learning scheme may also be applied to other NLP problems where labels may contain noise (\eg, obtained via crowdsourcing from non-experts); the augmentation and self-training method may be helpful for other settings like semi-supervised or few-shot learning.
One may also consider exploring larger pre-trained language models (\eg, RoBERTa-large) or more recent pre-trained language models (\eg, COCO-LM~\cite{meng2021coco}) for the distantly-supervised NER task.

\section*{Acknowledgments}
Research was supported in part by US DARPA KAIROS Program No.\ FA8750-19-2-1004, SocialSim Program No.\ W911NF-17-C-0099, and INCAS Program No.\ HR001121C0165, National Science Foundation IIS-19-56151, IIS-17-41317, and IIS 17-04532, and the Molecule Maker Lab Institute: An AI Research Institutes program supported by NSF under Award No.\ 2019897. Any opinions, findings, and conclusions or recommendations expressed herein are those of the authors and do not necessarily represent the views, either expressed or implied, of DARPA or the U.S. Government.
We thank anonymous reviewers for valuable and insightful feedback.

\bibliography{ref}
\bibliographystyle{acl_natbib}

\appendix

% \clearpage

\section{Generalized Cross Entropy}
\label{sec:gce}
The Generalized Cross Entropy (GCE) loss (Eq.~\eqref{eq:gce_loss}, also shown below),
\begin{equation*}
\mathcal{L}_{\text{GCE}} = \sum_{i=1}^n \frac{1 - f_{i, y_i}(\bs{x};\bs{\theta})^{q}}{q},
\end{equation*}
is a generalized version of the cross entropy (CE) loss, as $q \to 0$, $\mathcal{L}_{\text{GCE}} \to \mathcal{L}_{\text{CE}}$, shown as follows:
\begin{align*}
\lim_{q \to 0} \mathcal{L}_{\text{GCE}} &= \lim_{q \to 0} \sum_{i=1}^n \frac{1 - f_{i, y_i}(\bs{x};\bs{\theta})^{q}}{q}\\
&= \lim_{q \to 0} \sum_{i=1}^n \frac{\frac{d}{dq}\left( 1 - f_{i, y_i}(\bs{x};\bs{\theta})^{q} \right)}{\frac{d}{dq}(q)}\\
&= \lim_{q \to 0} \sum_{i=1}^n \frac{-f_{i, y_i}(\bs{x};\bs{\theta})^{q} \log f_{i, y_i}(\bs{x};\bs{\theta}) }{1} \\
&= - \sum_{i=1}^n \log f_{i, y_i}(\bs{x};\bs{\theta}) \\
&= \mathcal{L}_{\text{CE}}.
\end{align*}
The second line is obtained by applying L’Hôpital’s rule.

\section{Baseline Sources}
\label{sec:baselines}
We use the following sources for baseline implementation:
\begin{itemize}[leftmargin=*]
\item \textbf{Distant RoBERTa}: 
We use the Huggingface Transformer library for the RoBERTa model: \url{https://huggingface.co/transformers/}.
\item \textbf{AutoNER}: 
We use the authors' released code: \url{https://github.com/shangjingbo1226/AutoNER}.
\item \textbf{BOND}: 
We use the authors' released code: \url{https://github.com/cliang1453/BOND/}.
\end{itemize}
The results reported in Table~\ref{tab:results} are obtained by taking the higher value of (1) our own run and (2) the corresponding performance reported in \cite{liang2020bond}.

\section{Implementation Details}
\label{sec:impl}
\paragraph{Tagging Scheme for Distantly-Supervised NER.}
Instead of using BIO/BIOES tagging scheme, we use the binary IO format (\ie, only distinguish whether a token is a part of an entity or not) following previous work~\cite{peng2019distantly}, mainly because the distant labeling process may induce partially matched entities (\eg, the first sentence in Figure~\ref{fig:example}), and the beginning/ending token of the entity can be inaccurate.

\paragraph{Dropping Non-Entity Tokens From Distant Labels.}
We find it beneficial to randomly exclude a portion of distantly-labeled non-entity tokens (we dropped $50\%$ non-entity tokens for all three datasets in the experiments) from training. This is probably because the distant labeling process fails to detect some entities which will be mislabeled as the ``O'' class, and randomly dropping non-entity tokens reduces the number of such false negative labels.

\paragraph{Implementation of the NER Head.}
Different from the common setup of fine-tuning  PLMs for NER, we implement the NER head in Figure~\ref{fig:aug} as two linear layers instead of one: One linear layer classifies entity tokens against non-entity ones (\ie, binary classification), and the other linear layer classifies all entity type classes. During the self-training step, the former is trained to maintain its predictions, while the latter is trained to approximate the soft labels.

\paragraph{Noisy Label Removal for Minority Types.}
Some minority types may have very few labeled tokens and the model will output low-confidence predictions on them due to insufficient training. To make sure those tokens are not treated as noisy ones and removed from training, we do not perform noisy label removal on entity type classes (\ie, keep all tokens labeled as those classes) of which $>90\%$ tokens are predicted with confidence lower than the threshold.

\end{document}